# Bearing Fault Diagnosis using Graph Sampling and Aggregation Network

Jiaying Chen[1], Xusheng Du, Yurong Qian, Gwanggil Jeon

*Abstract*—Bearing fault diagnosis technology has a wide range of practical applications in industrial production, energy and other fields. Timely and accurate detection of bearing faults plays an important role in preventing catastrophic accidents and ensuring product quality. Traditional signal analysis techniques and deep learning-based fault detection algorithms do not take into account the intricate correlation between signals, making it difficult to further improve detection accuracy. To address this problem, we introduced Graph Sampling and Aggregation (GraphSAGE) network and proposed GraphSAGE-based Bearing fault Diagnosis (GSABFD) algorithm. The original vibration signal is firstly sliced through a fixed size non-overlapping sliding window, and the sliced data is feature transformed using signal analysis methods; then correlations are constructed for the transformed vibration signal and further transformed into vertices in the graph; then the GraphSAGE network is used for training; finally the fault level of object is calculated in the output layer of the network. The proposed algorithm is compared with five advanced algorithms in a real-world public dataset for experiments, and the results show that the GSABFD algorithm improves the AUC value by 5% compared with the next best algorithm.

## I. INTRODUCTION

BEARING fault diagnosis has a wide range of applications in aerospace [1], automotive [2], industrial manufacturing [3], railway [4], energy [5] and other fields. Bearing fault diagnosis is a topic of great importance with numerous practical applications. In-depth research on bearing fault diagnosis technology can effectively contribute to realizing the goals of preventive maintenance of machinery and equipment [6], preventing catastrophic accidents [7], prolonging the service life of equipment [8], and reducing the failure rate of machinery production products [9].

Traditional bearing fault detection methods, such as

This research was supported by Natural Science Foundation of Xinjiang Uygur Autonomous Region of China (2022D01C692), The Key Research and Development Project in Xinjiang Uygul Autonomous Region (No.2022B01006), Basic Research Foundation of Universities in the Xinjiang Uygur Autonomous Region of China (XJEDU2023P012), National Natural Science Foundation of China (62266043), Tianshan Innovation Team Program of Xinjiang Uygur Autonomous Region of China (2023D14012).(Corresponding author: Jiaying Chen.)

Jiaying Chen, and Yurong Qian are with the School of Software, Xinjiang University, Urumqi 830091, China (e-mail: chenjy@xju.edu.cn.).

Xusheng Du is with the School of Information Science and Eng ineering, Xinjiang University, Urumqi 830046, China (e-mail: duxus heng@stu.xju.edu.cn.)

Gwanggil Jeon is with the College of Information and Technolo gy, Incheon National University, Korea (gjeon@inu.ac.kr)

vibration analysis, acoustic detection, thermal detection and optical detection, have various limitations [10]. For example, these methods require on-site inspection and analysis by professional technical personnel, which can lead to subjectivity issues and difficulty in obtaining quantitative descriptions of fault severity. In addition, traditional methods are difficult to monitor failures in real time. Therefore, once a fault is detected, it can take a significant amount of time to analyze the cause, which can delay the repair time and affect the normal operation of the equipment. In addition, from a technical point of view, the accuracy of traditional fault diagnosis technology is not high, making it difficult to achieve rapid detection and localization of faults. These shortcomings lead to a decrease in the efficiency of bearing fault detection and diagnosis, and limit the reliability of bearings in equipment [11].

In recent years, bearing fault diagnosis technology based on deep learning has become a mainstream research direction. Compared with traditional methods, deep learning-based techniques can solve the problem of subjectivity of fault description, achieve quantitative description of fault damage degree, reduce the dependence on manual labor, monitor the bearing condition in real time, and improve the detection accuracy [12]. Currently, the mainstream deep learning-based methods include those based on convolutional neural networks, long- and short-term memory networks, generative adversarial networks, and autoencoders [13]. However, all of these methods assume that the objects to be detected are independently distributed, ignoring the correlation and complex topological relationships between objects. As a result, the information about the neighbors of the objects cannot be learned and used during the training process.

To address the problem of neglecting the intricate correlation between objects and their neighbors in traditional deep learning-based error detection methods, this paper presents a graph sampling and aggregation network-based storage error detection approach. This method can simultaneously learn the feature information of the object and its neighbors, and exhibits excellent scalability and generalization ability. The GraphSAGE-based bearing fault detection method maximizes the feature learning ability of the classical deep learning model, while also taking into account the correlation between objects and their neighbors, which further improves the accuracy of fault detection.

The main contributions of this paper can be briefly summarized as follows:

1. This paper presents a novel approach to convert intricate continuous vibration signals into graph nodes. The method involves extracting features from the original signal and consolidating the complex continuous data into a single object. Then, by taking into account the intricate correlations between objects, the original vibration signal is transformed into a graph.

2. In this paper, we improve Graph Sampling and Aggregation Network so that it can learn the correlation between vibration signals and improve the accuracy of bearing fault detection.

3. We validate the effectiveness of the proposed algorithm on real-world datasets using various metrics. Experiments show that the proposed algorithm significantly improves the detection accuracy of the algorithm after aggregating and learning the features of objects and their neighbors.

## II. RELATED WORK

As a ubiquitous mechanical component, bearings are critical to mechanical equipment because they support and transmit loads, reduce friction, and ensure equipment accuracy. Specifically, bearings are responsible for supporting the load of rotating parts and transmitting the load to other components. In addition, by incorporating steel balls and other components, bearings can significantly reduce mechanical losses during rotation. In addition, high precision bearings can significantly improve the accuracy of equipment [14].

Since bearings play a very important role in machinery and equipment, ensuring their proper operation and detecting bearing failures in a timely manner is a common concern for both academia and industry. A faster and more accurate method of diagnosing bearing failures can prevent catastrophic equipment accidents and reduce the production risk of companies [15].

The process of bearing fault diagnosis usually consists of three steps: data acquisition, feature extraction and health condition identification [16]. Traditional bearing fault detection methods include vibration signal-based, acoustic signal-based, temperature-based, etc. [17]. Among them, vibration signal is the most widely studied. Since the vibration signal contains rich information about the operating condition of the equipment and can obtain similar experimental results in multiple tests, the vibration signal-based detection method is one of the commonly used bearing fault detection methods. Qin et al [18] constructed a novel impulse wavelet based on the impulse vibration model generated by bearing failures, and then performed bearing failure diagnosis by extracting the impulse features. However, their method requires a large amount of computational resources to calculate the impulse features. Namdar et al [19] proposed a Kalman filter (KF) based SITF detection algorithm for mechanical fault detection, which extracts and discriminates rolling bearing faults by using statistical features and signal processing techniques. The shortcoming of this method is that the detection accuracy of the algorithm depends on the feature selection, which may be poor if inappropriate features are selected. Cheng et al [20] proposed an Adaptive Periodic Mode Decomposition (APMD) method to extract the periodic component (PC) without setting any parameters from the perspective of period identification and extraction. However, the method is ineffective in early fault detection, while valuable information is easily lost during the execution of the algorithm. Guo et al [21] proposed an improved empirical modal decomposition (EMD) method based on multi-objective optimization to extract the inner and outer ring failure characteristics of the bearing separately. The proposed approach can effectively reduce the number of IMFs and improve the accuracy of EMD. The disadvantage of this method is that it does not discriminate between different types of defects and the detection time is long.

Since traditional bearing fault diagnosis methods are mainly manual in terms of feature extraction and rely heavily on expert experience, modern industrial applications prefer to achieve autonomous fault diagnosis with the help of artificial intelligence and hope to improve the accuracy of diagnosis [22]. Machine learning is a branch of artificial intelligence, and fault diagnosis based on traditional machine learning methods mainly includes algorithms based on K-nearest neighbors (KNN) [23], singular value decomposition (SVD) [24], and so on. Konar et al [25] used support vector machine (SVM) together with continuous wavelet transform (CWT), an advanced signal processing tool, to analyze the frame vibrations during startup. However, the SVM and CWT-based methods may be more subjective in the feature selection process, resulting in less accurate detection results. Deng et al [26] introduced empirical modal decomposition, fuzzy information entropy, improved particle swarm optimization algorithm and least squares support vector machine in a fault diagnosis system to provide a new method for engine bearing fault diagnosis. However, their method requires a high level of manually set parameters, and inappropriate parameters can lead to different decomposition results and affect the detection performance of the algorithm. Toma et al [27] proposed a hybrid motor current data-driven method for bearing fault diagnosis using statistical features, genetic algorithm (GA) and machine learning models. Although their proposed method has a high detection accuracy, the model used in the method is more complex, which affects the detection efficiency.

Deep learning based bearing fault detection technique is able to extract its deep features from the input data by constructing a multilayer nonlinear network structure, which improves the detection accuracy and therefore is widely used [28]. Compared with other methods, deep learning-based methods do not require human involvement in feature extraction and selection during the detection process, while they are able to automatically detect the subtle differences between normal and faulty data from massive data, and have higher detection accuracy. Deep learning-based methods are also significantly better than other types of algorithms in terms of generalization and automation. Based on these advantages, deep learning-based storage fault detection methods have become the most interesting area for researchers in recent years. Janssens et al [29] first used CNN for bearing fault detection in 2016. The authors process the vibration signal by using it as an input to the CNN. The features are automatically extracted using, for example, the multilayer convolutional kernel pooling of the CNN. However, the method uses a small data set during the training process, which results in its weak generalization ability. Eren L et al [30] built a fast and accurate bearing fault detection system using a one-dimensional convolutional neural network (CNN), which combines feature extraction and fault detection into a single learning unit. The shortcoming of this method is the limitation of its feature extraction capability. Due to the limitation of the convolutional kernel size, the method may not be able to effectively extract the relevant features in the vibration signal. Guo et al [31] proposed a generative adversarial network (GAN) based fault diagnosis method. The method uses GAN networks to generate a large number of potential faulty objects, and then uses the generated faulty objects together with real faulty objects as the input of the fault classifier to improve the detection accuracy. However, the method requires a large amount of fault data for training, which may affect the final detection performance of the model if the amount of data is insufficient.

## III. MODEL

In this section, we present a novel approach to bearing fault detection using the GSABFD algorithm. The proposed method is based on unsupervised learning and consists of four main components: (1) a data converter, (2) an attributed graph constructor, (3) a feature extractor, and (4) a fault diagnosis. Specifically, the GSABFD first preprocesses the data and extracts features from the vibration signals using the data converter. Next, the attributed graph constructor builds a graph structure with high topological similarity based on the similarity between objects. Then, the feature extractor learns the embedding of the constructed graph structure data and aggregates the node information to obtain an effective representation of the graph structure data. Finally, the fault diagnosis determines the fault degree of the nodes in the 2016. The authors process the vibration signal by using it as an input to the CNN. The features are automatically extracted using, for example, the multilayer convolutional kernel pooling of the CNN. However, the method uses a small data set during the training process, which results in its weak generalization ability. Eren L et al [30] built a fast and accurate bearing fault detection system using a one-dimensional convolutional neural network (CNN), which combines feature extraction and fault detection into a single learning unit. The shortcoming of this method is the limitation of its feature extraction capability. Due to the limitation of the convolutional kernel size, the method may not be able to effectively extract the relevant features in the vibration signal. Guo et al [31] proposed a generative adversarial network (GAN) based fault diagnosis method. The method uses GAN networks to generate a large number of potential faulty objects, and then uses the generated faulty objects together with real faulty objects as the input of the fault classifier to improve the detection accuracy. However, the method requires a large amount of fault data for training, which may affect the final detection performance of the model if the amount of data is insufficient. attribute graph based on the reconstruction error. The overall architecture of the proposed GSABFD method is shown in Figure 1.

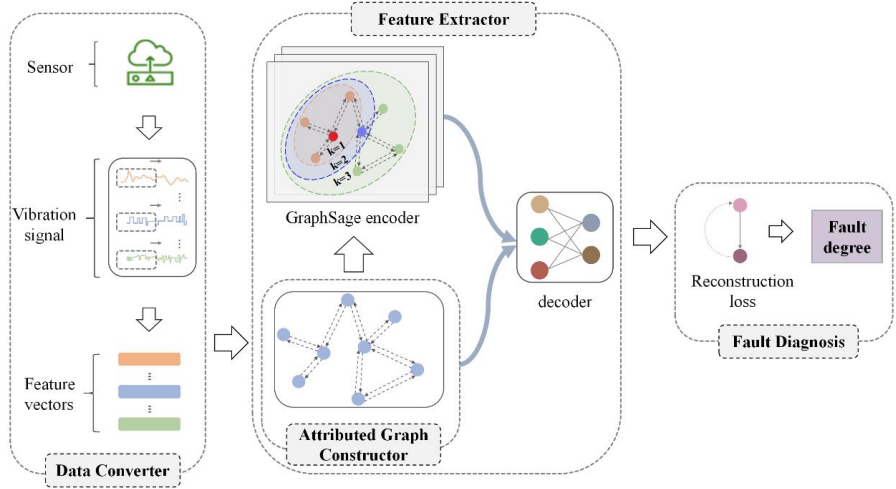

Figure 1. Overview of the GSABFD framework for bearing fault diagnosis. In this model, the vibration signal of the bearing is provided as input and the output produced by the model is the fault degree.

### A. Data Converter

The Data Converter is the first component of the GSABFD method and plays a crucial role in preprocessing and extracting features from the raw vibration signals. Its main purpose is to convert the vibration signal into a more discriminable feature representation. It is usually necessary to collect raw signal data from a variety of sensors installed on the machine. In bearing fault diagnosis studies, researchers typically rely on vibration signals for fault detection. Since the collected raw vibration signals typically contain a lot of redundant and irrelevant information, a manual feature extraction method is required to process the raw data and transform it into specific types of data required for downstream tasks to achieve highly accurate fault diagnosis. This allows the elimination of superfluous information while retaining relevant data for efficient and accurate fault detection.

To extract time-domain features, we specifically use nine time-domain metrics. Table 1 shows the formulas of the nine time-domain metrics, which are categorized into four dimensional parameters and five non-dimensional parameters. Nine time-domain feature values can be obtained using the following formula (see Equation (1)).

$$F_{time} = \left[ X_p, \sigma_X(t), \varphi_x^2(t), X_{rms}, I_p, C_f, S_f, k_1, k_3 \right] \quad (1)$$

Table 1. Time domain indicators

| mensional parameters | | non-dimensional parameters | |
|---|---|---|---|
| Peak | $X_p = max\ \|x_i(t)\|$ | Crest factor | $I_p = X_p/X_{rms}$ |
| Standard deviation | $\sigma_x(t) = \sqrt{\sum_{i=1}^{n}(x_i(t)-\mu_i(t))^2/(n-1)}$ | Impulse factor | $C_f = X_p/\|\bar{X}\|$ |
| mean square | $\varphi_x^2(t) = E[X^2(t)]$ | Shape factor | $S_f = X_{rms}/\|\bar{X}\|$ |
| Root mean square | $X_{rms} = \sqrt{\sum_{i=1}^{n}x_i^2(t)/n}$ | Kurtosis | $k_4 = \dfrac{E[(X-\mu)^4]}{\sigma^4}$ |
| | | Skewness | $k_3 = \dfrac{E[(X-\mu)^3]}{\sigma^3}$ |

In Equation (1), $X_p$ represents Peak, $\sigma_X(t)$ represents Standard deviation, $\varphi_x^2(t)$ represents mean square, $Xrms$ represents Root mean square, $I_p$ represents Crest factor, $C_f$ represents Impulse factor, $S_f$ represents Shape factor, $k_4$ represents Kurtosis, $k_3$ represents Skewness.

In this study, we use two methods to extract frequency features: the Daubechies Wavelet Transform and the Ensemble Empirical Mode Decomposition (EEMD). The Daubechies wavelet transform is a wavelet with a hierarchical property based on orthogonal wavelets. The classification of the Daubechies wavelet is based on the value of the vanishing moment, called the tap (also known as the number of vanishing moments). Typically, the length of the Daubechies wavelet is twice the length of the tap, and the filter is usually described as DN by the length of the filter $N$. In our investigation, we utilize the Daubechies wavelet as D20, and eight frequency features can be obtained through the Daubechies wavelet transform using the following formula (see Equation (2)):

$$F_{DB} = [DB1, DB2, DB3, DB4, DB5, DB6, DB7, DB8] \quad (2)$$

EEMD is an enhancement to Empirical Mode Decomposition (EMD) that addresses the issue of mode mixing in the Intrinsic Mode Functions (IMFs) obtained from EMD decomposition by using noise-assisted auxiliary signals. EEMD also improves the decomposition results by reducing the influence of boundary effects. The EEMD technique works by introducing normally distributed white noise into the signal to be analyzed and then performing EMD decomposition on the combined signal. This process is repeated several times with additional noise added to the signal until the noise completely covers the signal. Since the white noise spectrum is uniformly distributed, it will counterbalance itself after several averaging calculations. The final result is obtained by integrating and averaging the IMFs obtained from each iteration, calculating the EEMD yields the six features in Equation 3.

$$F_{EEMD} = [EEMD1, EEMD2, EEMD3, EEMD4, EEMD5, EEMD6] \quad (3)$$

The Data Converter extracts a set of 23 features, consisting of nine time-domain features, eight Daubechies wavelet features, and six EEMD features. The extracted features are highly discriminative and reduce the redundancy of the raw vibration signals. In summary, the extracted features can be expressed by the following formula (see Equation (4)):

$$Features = [F_{time}, F_{DB}, F_{EEMD}] \quad (4)$$

We present the pseudo-code of the Data Converter in Algorithm 1.

---

**Algorithm 1** Data Converter

**Input:** vibration signals: $X'$

**Output:** extracted features: $X$

1: Calculate the time domain characteristics of $X'$

2: $F_{time} = [X_p, \sigma_X(t), \varphi_x^2(t), X_{rms}, I_p, C_f, S_f, k_4, k_3]$

3: Calculate the frequency domain characteristics using Daubechies wavelet transform of $X'$

4: $F_{DB} = [DB1, DB2, DB3, DB4, DB5, DB6, DB7, DB8]$

5: Calculate the frequency domain characteristics using EEMD of $X'$

6: $F_{EEMD} = [EEMD1, EEMD2, EEMD3, EEMD4, EEMD5, EEMD6]$

7: $X = Concat([F_{time}, F_{DB}, F_{EEMD}], dim = 1)$

8: **Return** $X$

---

### B. Attributed Graph Constructor

The Attributed Graph Constructor aims to convert the data processed by the Data Converter into a directional attribute graph $G = (V, E)$, providing a more comprehensive and informative representation for further analysis and processing. Each data point in the sample set $G = (V, E)$ is represented by a vector $V_i$ corresponding to a node in the attribute graph $G$. The attribute of the node is represented by a group of $vectors\ (v_1, v_2, v_3, \cdots, v_m)$. The directional edges in the attribute graph connect the target node with its $k$ most similar neighbors. Specifically, the construction of the attribute graph is divided into four processes, which are described in detail below. The process of the Attributed Graph Constructor is illustrated in Figure 2.

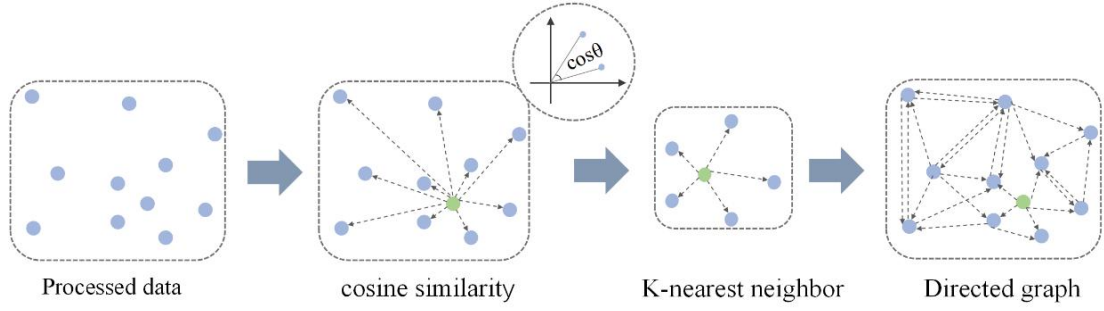

Figure 2 Flowchart of the Attributed Graph Constructor. For a clearer description, we default the value of k in the graph to 5.

(i) Node similarity calculation: To transform the initial Euclidean data into a graph structure, it is necessary to establish the connection relationships between the nodes. In general, there are two methods to evaluate the similarity between nodes: Euclidean distance and Cosine similarity. Due to the sensitivity of the Euclidean distance to the size of attribute and the fluctuation of the value range according to the dimensionality, and since fault diagnosis is more concerned with the relative differences between the data, we choose to use the Cosine similarity as a uniform measure to calculate the similarity between nodes. The similarity between nodes can be obtained using the following formula (see Equation (5)). In Equation (5), $v_i$, $v_j$ represent nodes in the attribute graph, and $x_{im}$, $x_{jm}$ represent the attributes of $v_i$ and $v_j$ respectively.

$$simility\left(v_i, v_j\right) = \frac{x_{i1} \cdot x_{j1} + x_{i2} \cdot x_{j2} + \cdots + x_{im} \cdot x_{jm}}{\sqrt{x_{i1}^2 + x_{i2}^2 + \cdots + x_{im}^2} \times \sqrt{x_{j1}^2 + x_{j2}^2 + \cdots + x_{jm}^2}} \quad (5)$$

(ii) K-nearest neighbor selection: The k most similar nodes to the target node are selected as its neighbors, denoted as $N_k\left(v_i\right)$. This selection is based on the Cosine similarity previously calculated between the target node and all other nodes in the graph. The k-nearest neighbor can be obtained by the following formula (see Equation (6)). In Equation (6), $N_k\left(v_i\right)$ represents the k most similar nodes to $v_i$, and $\left(v_1, v_2, \cdots, v_m\right)$ represent nodes in the attribute graph.

$$N_k\left(v_i\right) = \max\_simility_k\left(v_1, v_2, ..., v_m\right) \quad (6)$$

(iii) Directional graph construction: To provide a more comprehensive explanation of how we construct the attribute graph, we assign weight values to the directed edges based on the degree of similarity between the nodes. The weight of each edge falls between 0 and 1, and is proportional to the similarity score between the target node and its neighbors. To ensure that the sum of weights of all edges from the target node to its neighbors is 1, we normalize the similarity scores of all neighbor nodes. Hence, a larger weight value between two nodes denotes a higher degree of similarity. On the other hand, the weight of edges from the target node to non-neighboring nodes is assigned a value of 0. The weights of edges in the attribute graph are calculated using the following formula (see equation (7)). In Equation (7), $N_k\left(v_i\right)$ represents the k most similar nodes to $v_i$, and $v_i$, $v_j$ represent nodes in the attribute graph.

$$e_{i,j} = \begin{cases} simility\left(v_i, v_j\right) / \sum_1^k simility\left(v_i, v_k\right) & if \ x_j \in N_k\left(v_i\right) \\ 0 & otherwise \end{cases} \quad (7)$$

(iv) Graph initialization: The graph information is typically represented by an adjacency matrix, denoted as $matrix_{m \times n}\left[e_{i,j}\right]$. In our model, each learning iteration for the target node requires not only the information of its surrounding neighbors, but also its own attributes. However, the matrix representation of the directed graph constructed based on the neighborhood information alone does not integrate the attributes of the node itself. Therefore, we need to perform an initialization operation by adding an identity matrix $E_m$ of the same size as the data set to the original adjacency matrix in order to incorporate the node attributes. After initialization, the matrix **M** can be represented by the following formula (see Equation (8)). In Equation (8), $matrix_{m \times n}\left[e_{i,j}\right]$ represents the adjacency matrix of the graph, and $E_m$ is an identity matrix.

$$M = matrix_{m \times n}\left[e_{i,j}\right] + E_m \quad (8)$$

We present the pseudo-code of the Attributed Graph Constructor in Algorithm 2.

---

**Algorithm 2** Attributed Graph Constructor

**Input:** $X\left(v_1, v_2, v_3, \cdots, v_m\right) \in R^{m \times 23}$ (Output of Algorithm 1)

**Output:** $M$ (Adjacency matrix)

1: **while** i < m **do**
2:   **for** j in m:
3:     $simility\left(v_i, v_j\right)$
4:     $N_k\left(v_i\right) = \max\_simility_k\left(v_1, v_2, \cdots, v_n\right)$
5:     $e_{i,j} = \begin{cases} simility\left(v_i, v_j\right) / \sum_1^k simility\left(v_i, v_k\right) & if \ x_j \in N_k\left(v_i\right) \\ 0 & otherwise \end{cases}$
6:     $M = matrix_{m \times n}\left[e_{i,j}\right] + E_m$
7: **return** $M$

---

### C. Feature Extractor

The feature extractor is a crucial component of the model as it extracts essential features from the raw data and transforms them into higher-level feature vectors. These vectors capture the most informative and representative aspects of the input data, making them more suitable for subsequent processing and analysis. In real-world scenarios, unknown fault information may be present, which is why our model uses an unsupervised learning approach. To achieve this, we employ an autoencoder, a neural network model that is commonly used for unsupervised learning tasks, as a feature extractor to capture intrinsic data features. However, traditional autoencoders are primarily designed for structured data, and their ability to process unstructured data remains unsatisfactory.

To better handle graph-structured data and provide stronger feature learning capabilities, we use GraphSAGE to build an encoder capable of learning contextual information of nodes in a graph. GraphSAGE generates representation vectors of nodes by aggregating feature vectors of their neighborhoods, enabling it to better handle graph-structured data and provide stronger feature learning capabilities compared to traditional null-domain convolutional neural networks.

Figure 3 illustrates the feature extractor process. After passing through the graph constructor, the data enters the GraphSAGE encoder for sampling, aggregation, and updating. The aggregated feature vector data is then reconstructed in the decoder, which is constructed using a fully connected network. The decoder decodes the encoded feature vectors back to the original data representation using inverse operations to ensure accurate reconstruction. This allows us to maintain data integrity throughout the feature extraction process.

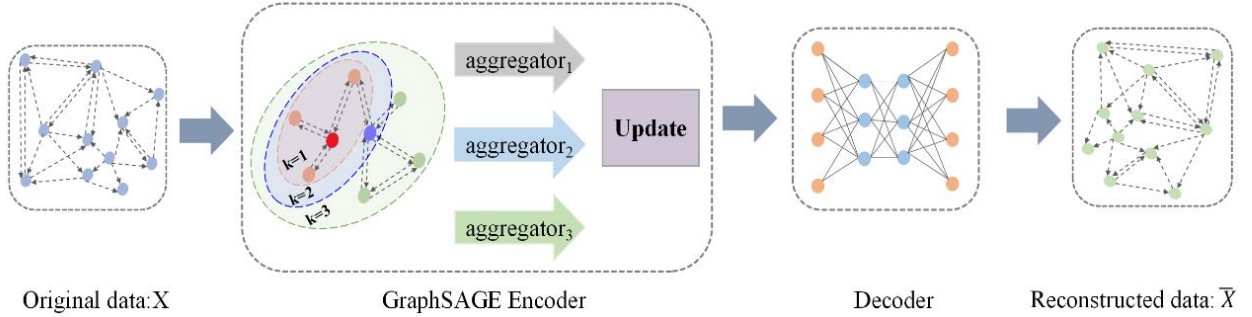

Original data:X    GraphSAGE Encoder    Decoder    Reconstructed data: $\bar{X}$

Figure 3. Flowchart of the Feature Extractor.

The key part of the feature extractor is GraphSAGE, which learns an aggregation function that effectively uses the information of neighboring nodes to learn the embedding of the target node, and updates the node information using an update function. The steps of the Feature Extractor are as follows:

(1) Sampling Neighbors of Nodes: Not all neighbors are equally informative, so we use random sampling of a subset of neighbors to reduce computational complexity. To increase efficiency, we sample an equal number of k-order neighbors for all nodes in the graph.

(2) Aggregating Information from Neighboring Nodes: We aggregate information from *k*-order neighbors to generate embedding's of *k-1* order neighbors, followed by aggregating information from *k-1* order neighbors to produce embedding's of *k-2* order neighbors. This iterative process continues until the target node is reached. As the number of iterations increases, the aggregated information of each node becomes increasingly global. The aggregation formula is as follows (see Equation (9)):

$$h_v^k \leftarrow \sigma\left(W \cdot MEAN\left\{h_v^{k-1}\right\} \cup \left\{h_u^{k-1}, \forall u \in N(v)\right\}\right) \quad (9)$$

In Equation (9), $N(v)$ represents the set of neighboring nodes of v in the graph, $h_v^k$ represents the k-th iteration of the hidden state vector for node $v$ in graph neural network. $h_u^{k-1}$ represents the hidden state vector of node $u$ in the *(k-1)-th* iteration of the network, it is used to capture information from the neighboring nodes of *v*. *W* represents the weight matrix that is learned during training, and $\delta$ is the activation function.

(3) Updating Node Representations: We use a neural network to update the representation vector of each node by combining its own features with the aggregated features of its neighbors, resulting in embedding for all vertices.

(4) Reconstructing Data: The aggregated feature vectors are fed into a decoder built from fully connected networks to reconstruct the original data representation. To ensure accurate reconstruction of the encoded feature vectors back to their original data, we use inverse operations to decode them.

We present the pseudo-code of the Feature Extractor in Algorithm 3.

---

**Algorithm 3** Feature Extractor

**Input:** $M$ (Output of Algorithm 2), $X\left(v_1, v_2, v_3, \cdots, v_m\right)$ (Output of Algorithm 1), epoch

**Output:** $\bar{X}$

1: $h_v^0 \leftarrow v_m$

2: **while** i < epoch **do**

3:     **for** k=0 to n do:

4:         **for** v∈ X do:

5:             $h_{N(v)}^k \leftarrow \sigma\left(W \cdot MEAN\left\{h_v^{k-1}\right\} \cup \left\{h_u^{k-1}, \forall u \in N(v)\right\}\right)$

6:             $h_v^k \leftarrow \sigma\left(W * CONCAT\left(h_v^{k-1}, h_{N(v)}^k\right)\right)$

7:         **end for**

8:     **end for**

9:     $Z \leftarrow h_v^k / \| h_v^k \|_2$

10:     $\bar{X} \leftarrow decoder(Z)$

11:     $loss \leftarrow loss\_fn\left(\bar{X}, X\right)$

12:     $loss.backward()$

13: **Return** $\bar{X}$

---

## D. Fault Diagnosis

After training the Graph Sample and Aggregate Network, we obtain the embedding of the original graph: $Z$, which is then fed to the decoder for reconstruction to obtain $\overline{X}$. Finally, we compare the reconstructed graph data $\overline{X}$ with the original graph data $X$ and calculate the reconstruction error of each node in the graph. The higher the reconstruction error, the more likely it is that the current node is faulty data. In other words, the reconstruction error of the faulty sample is greater. Therefore, we take the reconstruction error of each node as the final fault degree. The corresponding calculation formula is as follows (see Equation (10)):

$$FaultDegree = \frac{1}{2}\left(X - \overline{X}\right)^2 \tag{10}$$

In Equation (10), $X$ represents the original graph data, and $\overline{X}$ represents the reconstructed graph data. The reason why the reconstruction error can accurately identify the cause of faulty bearings is because during the model training process, the number of abnormal data is smaller than the normal data. Therefore, to reduce the loss, the feature extractor will prioritize fitting the normal data. As the number of iterations increases, the fit of the model to the normal data will improve, and the embedding information of the obtained nodes will be more biased towards the structure of the normal data. However, the information of faulty bearings that experience failures differs significantly from that of normal bearings. Consequently, when the embedded representation of faulty bearings is input to the decoder for reconstruction, the reconstructed data will be significantly different from the original data. Therefore, the magnitude of the reconstruction error can be used to determine whether the corresponding bearing is faulty. The fault diagnosis process is illustrated in Figure 4.

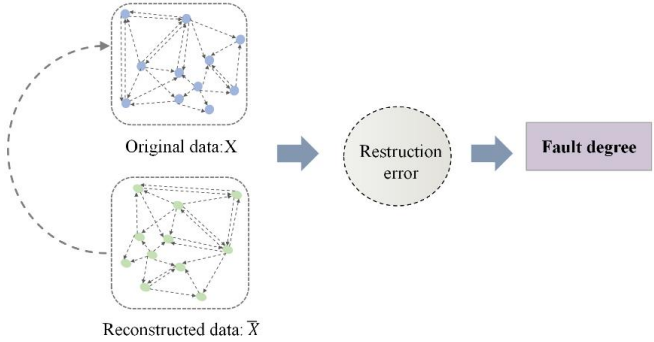

Figure 4. Flowchart of the Fault Diagnosis.

## IV. EXPERIMENTS

### A. Dataset

#### 1) Dataset and experimental environment introduction

The Case Western Reserve University Bearing Data is a publicly available dataset for the development and comparison of bearing fault diagnosis algorithms, acquired in Canada in 1994 by Isaiah K. Lloyd et al [32]. The vibration signals of the machine were collected in the radial and axial directions of the bearing using multiple sensors during the acquisition process. The collected vibrations consisted of normal, inner race fault, outer race fault and ball fault with a sampling frequency of 12 KHz and were collected at one second intervals.

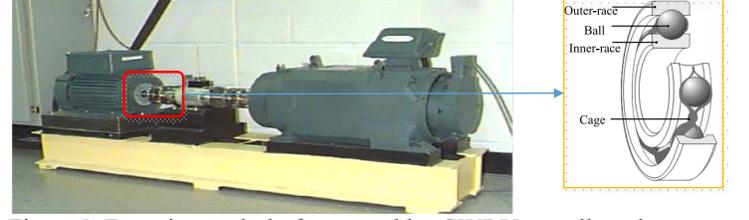

Figure5. Experimental platform used by CWRU to collect data

In this paper, we use a server with an Intel (R) Xeon Gold 5117 CPU with 28 cores, 256 GB of RAM, 2 TB of SSD, and Windows Server 2019 Standard as the operating system. The code editing tool used for the experiment is Matlab 2019A.

#### 2) Dataset processing methods

In this experiment, the vibration signals collected from the drive side of the CWRU data set are used. Specifically, the normal signal of the drive side is a matrix of 243938*1, and the fault types are inner race fault, outer race fault and ball fault which are matrices of 120000*1 respectively. By using a sliding window of size 300, the original data is first sliced without overlapping. The slice diagram is as follows.

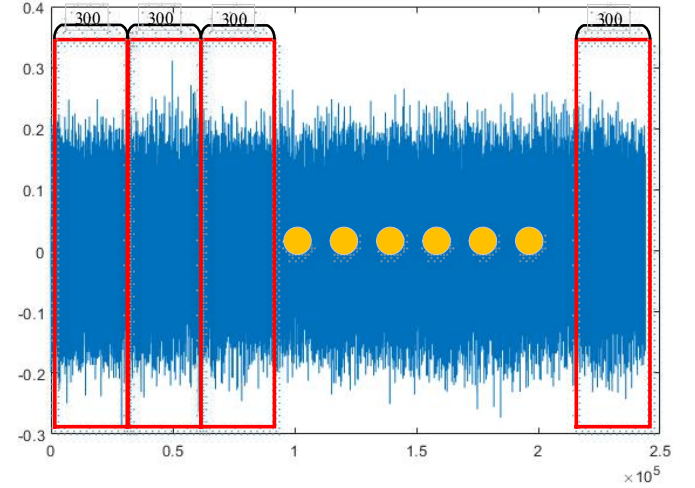

Figure 6. Slice the original dataset

The sliced dataset has a matrix of 300*800 for normal objects and 300*400 for each of the three types of faulty objects. Then the feature transformation is performed on the sliced dataset, and the transformed normal objects are 23*800, and the three types of faulty objects are 23*400 matrices respectively. Then the dataset is combined and 60 faulty objects are added in order among the 800 normal objects. In selecting the faulty objects, random sampling without replacement is used. The combined data set is the final data set. The combined inner circle data set to be tested is shown in Figure 7, where 0.1778, 0.3556, 0.5334 indicate the diameter of the fault.

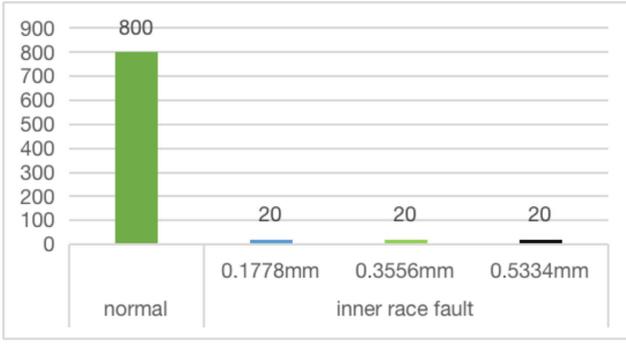

Figure 7. Example of the combined inner race fault to be detected

## B. Experiment design

### 1) Comparison algorithm introduction

In this paper, we convert the fault detection problem of bearings to anomaly detection in the field of machine learning. Therefore, in order to verify the effectiveness of the proposed algorithm, we select representative algorithms in the field of anomaly detection for comparison.

Each of these five algorithms has its own focus on anomaly detection, with some focusing on detecting global anomalous objects and others on detecting local anomalous objects. A comparison with these algorithms provides a more objective indication of the effectiveness of the proposed algorithms.

Table 2. Comparison algorithm

| Type of algorithm | Name of the algorithm |
|---|---|
| Generative adversarial net-based | SO-GAAL(Single Objective Generative Adversarial Active Learning) [33] |
| Autoencoder-based | AE(Autoencoder) [34] |
| Density-based | LOF(Local Outlier Factor) [35] |
| Distance-based | KNN(K-nearest neighbor) [36] |
| Isolation-based | IForest(Isolation Forest) [37] |

### 2) Evaluation Metrics

A single evaluation metric often does not objectively reflect the overall performance of the algorithm. Therefore, in our experiments, we choose AUC (Area under the Curve), ACC (Accuracy), DR (Detection Rate) and runtime as evaluation metrics. Higher values of AUC, ACC and DR indicate better performance of the algorithm. The shorter the running time, the more efficient the algorithm is in detecting.

$$AUC = \frac{\sum pred_{pos} > pred_{neg}}{positiveNum * engativeNum} \quad (11)$$

$$ACC = \frac{TP + TN}{TP + TN + FP + FN} \quad (12)$$

$$DR = \frac{TP}{TP + FN} \quad (13)$$

In Equation (11), predpos denotes the number of samples with positive prediction results, and predneg denotes the number of samples with negative prediction results. The denominator indicates the number of combinations of positive and negative samples. In Equation(12), (13), TP and TN are the correct prediction results of the model. FP and FN are the incorrect prediction results of the model.

### 3) Parameter Setting

In this experiment, we use a total of six algorithms. The type and number of parameters to be set for each algorithm varies, so in this section we describe the settings of each algorithm parameter in detail. The experiment was repeated 10 times under each parameter and the average value was used as the final performance evaluation of the algorithms.

(I) GSABFD, which contains two core parameters, the number of nearest neighbors $k$ and the sampling ratio. The parameter $k$ is set from 10 to 100 and the sampling ratio is set from 0.1 to 1; (II) SO-GAAL and AE both need to set the learning rate and the number of iterations, in the experiments, the learning rate is set to 0.001~0.01 and the number of iterations is set to 10~100; (III) The nearest neighbor parameter k of LOF and KNN is set to 10~100; (IV) The number of isolated trees in IForest is set to 256 by default, but the sampling size is set to 100~ 256.

## C. Experiment Results

In this section, we first show the detection results of the GSABFD algorithm; then we compare the results of the proposed algorithm with the comparison algorithm and analyze the reasons; finally, investigate the parameter sensitivity of the GSAFBD method.

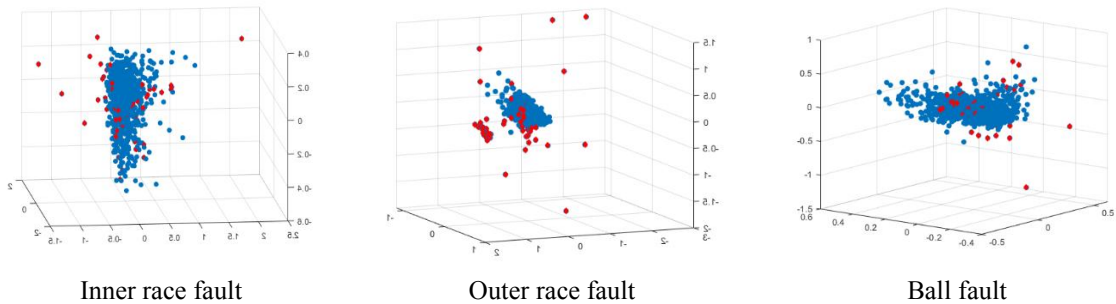

Inner race fault     Outer race fault     Ball fault

Figure 8. Visualization raw data

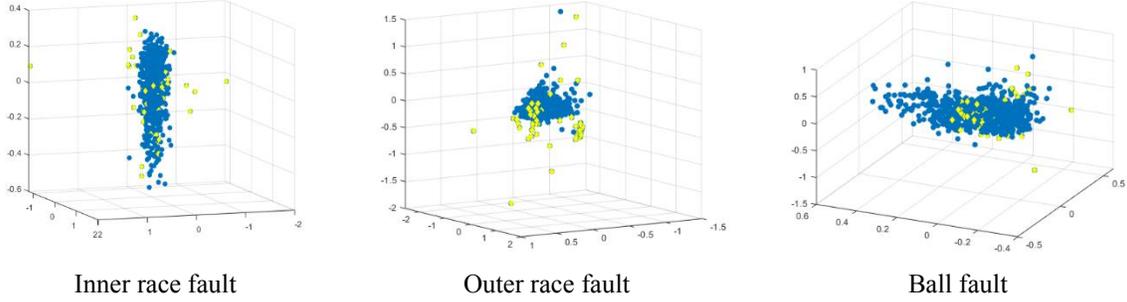

| Inner race fault | Outer race fault | Ball fault |

Figure 9. Detection result of GSABFD

By observing Figure 8, it is observed that certain faulty objects marked by the red dots exhibit high resemblance with normal objects, and some of the faulty objects are even mistakenly classified into the normal region. Many algorithms are only capable of detecting faulty objects that deviate significantly from the normal ones, while the accurate detection of faulty objects with a lower degree of deviation remains challenging. However, by observing the detection outcomes of the GSABFD algorithm shown in Figure 9, it is discovered that the algorithm demonstrates a high detection accuracy for faulty objects with a lower degree of deviation. This is because in the GraphSAGE network, the faulty objects aggregate the features of their neighboring nodes in each layer, which leads to a greater degree of deviation in the output layer of the network for the faulty objects that originally reside within the normal object region, and consequently, facilitates the successful detection of such faulty objects.

Table 3. Experimental results of six algorithms on four evaluation metrics

| AUC | | | | | | |
|---|---|---|---|---|---|---|
| Dataset | GSABFD | SO-GA AL | AE | LOF | KNN | IForest |
| Inner race fault | **0.99** | 0.89 | **0.93** | 0.71 | 0.69 | 0.92 |
| Outer race fault | **0.99** | 0.95 | 0.91 | 0.94 | 0.87 | **0.96** |
| Ball fault | **0.96** | 0.86 | 0.87 | 0.73 | 0.71 | **0.92** |

(I) AUC

| ACC (%) | | | | | | |
|---|---|---|---|---|---|---|
| Dataset | GSABFD | SO-G AAL | AE | LOF | KNN | IForest |
| Inner race fault | **99.53** | 96.51 | **97.20** | 93.72 | 93.02 | 97.21 |
| Outer race fault | **99.76** | 97.21 | 95.34 | 96.27 | 94.88 | **97.91** |
| Ball fault | **97.21** | 96.51 | 92.79 | 91.62 | 91.16 | **97.21** |

(II) ACC

| DR (%) | | | | | | |
|---|---|---|---|---|---|---|
| Dataset | GSABFD | SO-GA AL | AE | LOF | KNN | IForest |
| Inner race fault | **96.67** | 75 | **80** | 55 | 30 | 80 |
| Outer race fault | **98.33** | 80 | 66.67 | 73.33 | 63.33 | **85** |
| Ball fault | **80** | 75 | 48.33 | 40 | 36.67 | **80** |

(III) DR

| Runtime (second) | | | | | | |
|---|---|---|---|---|---|---|
| Dataset | GSABFD | SO-GA AL | AE | LOF | KN N | IForest |
| Inner race fault | 0.53 | 0.82 | **0.36** | 2.21 | 1.52 | 0.78 |
| Outer race fault | 0.59 | 0.94 | **0.33** | 2.35 | 1.23 | 0.92 |
| Ball fault | **0.62** | 1.11 | **0.29** | 2.42 | 1.37 | 0.88 |

(IV) Runtime

Observing the experimental results in Table 3, the following interesting conclusions can be found.

(i) The GSABFD algorithm achieves the best detection results in the first 3 metrics aimed at detection accuracy. Compared to the next highest IForest algorithm, it improves 5% of AUC value, 1.4% of ACC, and 12.24% of DR value. IForest is currently the most used anomaly detection algorithm in industry, and its excellent detection performance has been applied by most of the literature in the field to compare algorithms. The experiments demonstrate that the GSABFD algorithm outperforms IForest in detection performance and validates the effectiveness of the proposed algorithm.

(ii) Among the 3 types of faults, namely inner race fault, outer race fault and ball fault, the ball fault is the most difficult to detect. Most algorithms are significantly less accurate in detecting ball faults than the other two types of faults. This is because the cracks in the ball are smaller compared to the latter two, while the bearings generally contain multiple balls, and the fault caused by a certain ball may be similar to the normal balls, increasing the difficulty of detection.

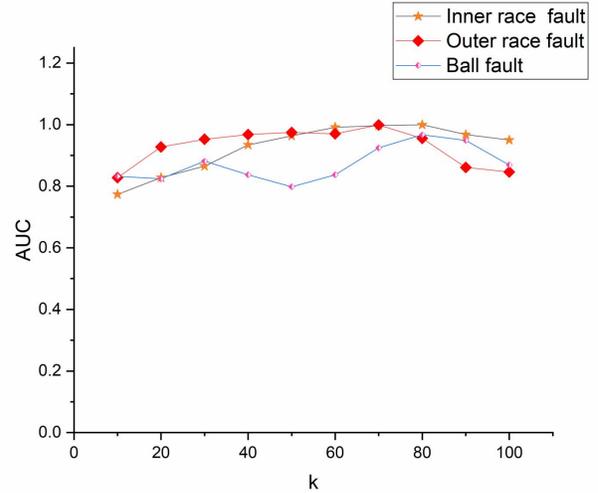

Figure 10. Effect of parameter k on the performance of GSABFD algorithm

Figure 10 shows the effect of different parameter $k$ values on the final detection performance of the proposed algorithm. As the $k$ value increases from small, the AUC value of GSABFD gradually increases and then starts to gradually decrease after reaching the apex. The main reason for this phenomenon is that when the $k$ value gradually increases from too small, the neighbor information gathered by each object starts to become more, and the feature information of the object learned by the GraphSAGE network becomes richer, increasing the detection performance of the algorithm. However, when the value of $k$ is too large, the phenomenon of over smoothing occurs due to too much information about the aggregated neighbors, which reduces the performance of the GSABFD algorithm.

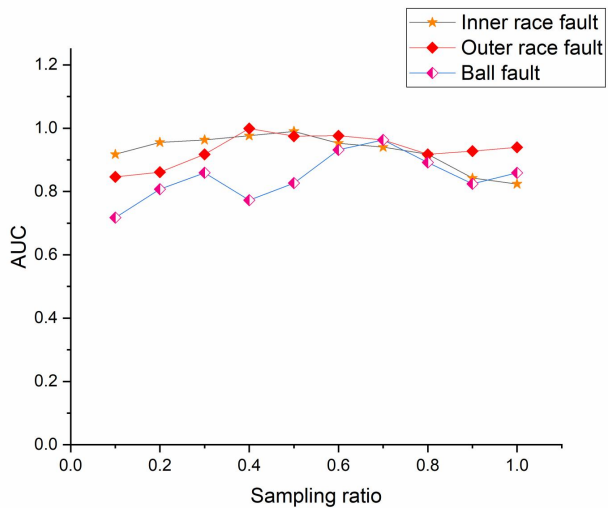

Figure 11. Effect of sampling ratio on the performance of GSABFD algorithm

The size of the sampling ratio directly affects the number of neighbor node information aggregated by each hidden layer of GraphSAGE. The higher the sampling ratio, the more neighbor information is aggregated. Observing Figure 11, we can see that the detection performance of GSABFD algorithm decreases from low to high as the sampling ratio changes from small to large. Combining Figure 10 and Figure 11, it can be seen that both $k$ value and sampling ratio affect the degree of aggregation of neighbor information by GSABFD algorithm. Setting the appropriate aggregation level can significantly improve the detection performance of the algorithm.

## V. CONCLUSION

In this paper, we determine the fault level of an object by jointly inputting the correlation between the object and its neighbors into the Graph Sampling and Aggregation network and calculating the reconstruction error of the object in the output layer of the network. The experimental results are compared with five representative anomaly detection algorithms on the CWRU dataset, and show that the GSABFD algorithm proposed in this paper can increase the deviation degree between normal and faulty objects and improve the detection accuracy of the algorithm after adding the calculation of correlation between objects. In the experiments of parameter sensitivity analysis, we found that the size of the sampling ratio and the number of nearest neighbors can cause different degrees of oversmoothing problems for the GSABFD algorithm. In future work, we will try to solve the problem of overfitting and information redundancy of the model due to aggregating too much information of neighbors, so that the depth of the network can be deepened while the feature information of neighbors can be fully utilized.

## ACKNOWLEDGMENTS

The authors are grateful to the anonymous referees for their insightful suggestions and comments. This research was supported by Natural Science Foundation of Xinjiang Uygur Autonomous Region of China (2022D01C692), The Key Research and Development Project in Xinjiang Uygul Autonomous Region (No.2022B01006), Basic Research Foundation of Universities in the Xinjiang Uygur Autonomous Region of China (XJEDU2023P012), National Natural Science Foundation of China (62266043), Tianshan Innovation Team Program of Xinjiang Uygur Autonomous Region of China (2023D14012).

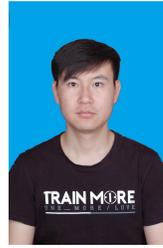

**Xusheng Du** was born in 1995. He is currently studying for his Ph.D. at the College of Information Science and Engineering, Xinjiang University. His main research interests include outlier detection, data mining, and deep learning.

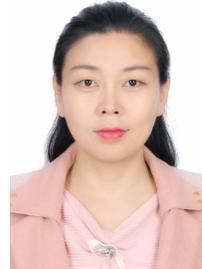

**Yurong Qian** received her B.S. and M.S. degree in the School of Information Science and Engineering, Xinjiang University, China, in 2002 and 2005 respectively, and Ph.D. in biology from Nanjing University, China in 2010. From 2012 to 2013, she worked as a postdoctoral fellow in the Department of Electronics and Computer Engineering, Hanyang University, South Korea. She is currently a professor in the School of Software, Xinjiang University, China. Her research interests include big data processing, image processing, as well as computational intelligence such as artificial neutral networks.

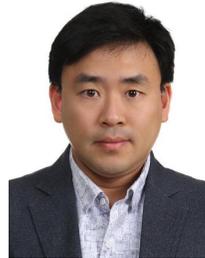

**Gwanggil Jeon** received his B.S. and M.S. degree in Electronics and Computer Engineering, Hanyang University, Korea, in 2008 and 2005, respectively. His main research interests focus on image processing, Video CODEC and computational intelligence.

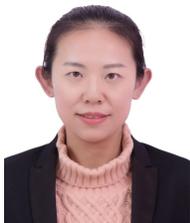

**Jiaying Chen** was born in 1988. She received the M.S. degree in software engineering and Ph.D. degree in technology of computer science from Xinjiang University, China, in 2017 and 2022 , respectively. Her main  research interests focus on machine learning, data mining, recommender system, etc.